\documentclass[manuscript,screen]{acmart}

\usepackage{algorithm,algpseudocode}
\usepackage{subfigure}
\usepackage{multirow}
\usepackage{soul}
\setcopyright{acmlicensed}
\copyrightyear{2024}
\acmYear{2024}
\acmDOI{XXXXXXX.XXXXXXX}

\begin{document}
\title{A Green Multi-Attribute Client Selection for Over-The-Air Federated Learning: A Grey-Wolf-Optimizer Approach}

\author{Maryam Ben Driss}
\affiliation{%
  \institution{Department of Computer Science, University of Quebec at Montreal, H2L 2C4}
  \city{Montreal}
  \country{Canada}  
}

\author{Essaid Sabir }
\affiliation{%
  \institution{Department of Science and Technology, TÉLUQ, University of Quebec, Montreal, H2S 3L4}
  \city{Montreal}
  \country{Canada}
}

\author{Halima Elbiaze}
\affiliation{%
  \institution{Department of Computer Science, University of Quebec at Montreal, H2L 2C4}
  \city{Montreal}
  \country{Canada}}

\author{Abdoulaye Baniré Diallo}
\affiliation{%
  \institution{Department of Computer Science, University of Quebec at Montreal, H2L 2C4}
  \city{Montreal}
  \country{Canada}}

\author{Mohamed Sadik}
\affiliation{%
  \institution{NEST Research Group, LRI Lab, ENSEM, Hassan II University of Casablanca}
  \city{Casablanca}
  \country{Morocco}}

\renewcommand{\shortauthors}{M. Ben Driss, et al.}
\begin{abstract}
  Federated Learning (FL) has gained attention across various industries for its capability to train machine learning models without centralizing sensitive data. While this approach offers significant benefits such as privacy preservation and decreased communication overhead, it presents several challenges, including deployment complexity and interoperability issues, particularly in heterogeneous scenarios or resource-constrained environments. Over-the-air (OTA) FL was introduced to tackle these challenges by disseminating model updates without necessitating direct device-to-device connections or centralized servers. However, OTA-FL brought forth limitations associated with heightened energy consumption and network latency. In this paper, we propose a multi-attribute client selection framework employing the grey wolf optimizer (GWO) to strategically control the number of participants in each round and optimize the OTA-FL process while considering accuracy, energy, delay, reliability, and fairness constraints of participating devices. We evaluate the performance of our multi-attribute client selection approach in terms of model loss minimization, convergence time reduction, and energy efficiency. In our experimental evaluation, we assessed and compared the performance of our approach against the existing state-of-the-art methods. Our results demonstrate that the proposed GWO-based client selection outperforms these baselines across various metrics. Specifically, our approach achieves a notable reduction in model loss, accelerates convergence time, and enhances energy efficiency while maintaining high fairness and reliability indicators.
\end{abstract}

\keywords{Over-The-Air Federated Learning; Client Selection; Grey Wolf Optimizer; Convergence Speed; Energy Efficiency; Reliability; Fairness.}

\ccsdesc[500]{Computing methodologies~Machine learning}

\maketitle

\section{Introduction}
\begin{figure}
\centering
\includegraphics[width=8.5cm]{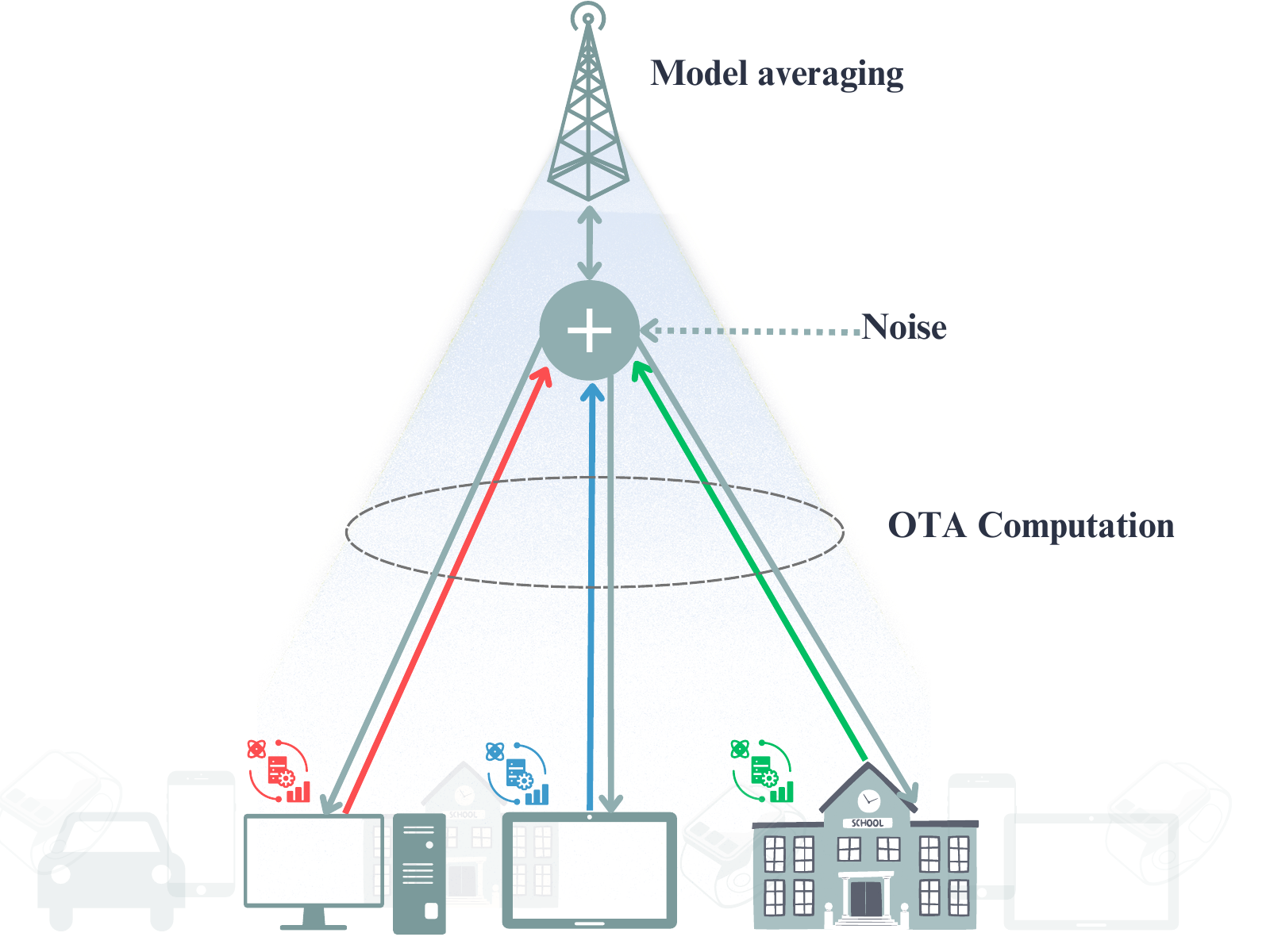}
\caption{Over-The-Air (OTA) federated learning process.}
\Description{.}
\label{FL}
\end{figure}

Artificial intelligence (AI) can transform many aspects of human society. With applications spanning healthcare, education, finance, transportation, and beyond, AI's capacity to analyze extensive datasets, predict outcomes, and automate tasks stands poised to enhance efficiency, accuracy, and the overall quality of life. However, traditional machine learning (ML) in massive and sensitive environments faces several challenges caused by the nature of large-scale datasets, distributed data sources, and their constraints such as data privacy, limited resources, and network heterogeneity. To address these issues, federated learning (FL) is a promising approach to train ML algorithms where devices collaborate to create and improve a shared model while preserving users' privacy and reducing communication overhead \cite{driss2023federated,luzon2024tutorial}. Instead of sending raw data to a central server for aggregation, each device maintains its dataset, trains a local model, sends model updates or gradients to the server that aggregates these updates, and then sends back the refined model to the individual devices. This process iterates until the global model reaches the desired accuracy. Over-the-air federated learning (OTA-FL) \cite{xiao2023over} is a specific implementation of FL that uses wireless communication channels for transmitting model updates as illustrated in Fig.\ref{FL} which greatly reduces the cost of communicating model updates from the edge devices.

Implementing OTA-FL in heterogeneous scenarios, where clients have different data distribution, limited bandwidth, and less reliable network conditions, faces several challenges including limited computing capabilities, data quality, and fairness between FL agents. Thus, the client selection step is crucial and the set of participants in each training round is a key factor in addressing these challenges and enhancing the learning process \cite{khan2020federated,azimi2024scalable}. By strategically choosing clients based on their data quality and computational capabilities, the FL system can effectively navigate through communication constraints, privacy concerns, and other challenges. Additionally, the client selection process is essential for ensuring the efficiency of model update distribution and the quality of the aggregated global model \cite{mayhoub2024review}. It ensures that the devices involved make valuable contributions to the shared learning, improving the overall effectiveness of FL algorithms. By carefully selecting which devices participate in the collaborative learning process, we can maximize the impact of each contribution, leading to better model performance and more reliable results.

In this paper, we introduce an optimization problem focused on crafting a multi-attribute client selection framework using the grey wolf algorithm. Our goal is to consider various criteria, including model accuracy, communication cost, resource reliability, and fairness among FL clients. This framework aims to strategically select clients based on these attributes, optimizing the overall performance and efficiency of the FL process. The remainder of this paper is organized as follows: Section II presents background and reviews previous research employing various techniques and criteria for client selection, outlining our novel contribution. Section III provides the system model and defines the objective function. Section IV elaborates on the GWO mechanism for client selection in the learning process. Section V presents experimental results and insights gleaned from our study. Finally, Section VI concludes the paper by summarizing key findings and suggesting avenues for future research.

\section{Background}
\subsection{Federated Learning}
FL is designed to train models across a network of decentralized devices while keeping data private. Individual devices collaboratively create a shared model without the need to share their raw data with a central server (see Fig.\ref{FL}). The process involves the following steps:
\begin{enumerate}
    \item \emph{Initialization:} A central server generates an initial global model for the selected devices.
    \item \emph{Local training:} Each device trains a local model using its dataset, which is not transmitted to the central server. 
    \item \emph{Local model update:} Each device produces a model update or gradient based on the differences between its local and global models.
    \item \emph{Aggregation:} The central server collects the updates from all participants and refines the global model through aggregation (e.g., averaging). 
    \item \emph{Global model update:} The refined global model is then sent back to all participating devices.
    \item \emph{Iteration:} The process of local training, local model update, aggregation, and global model update is repeated over multiple iterations. As each iteration progresses, the overall model's performance improves.
\end{enumerate}

OTA-FL is a promising concept that allows clients to share the same spectral resources by transmitting their local model updates and aggregating these models over the air in a "one-time" manner. This enables the efficient sharing of model updates over the air without relying on traditional wired networks \cite{zhu2024over}. The benefits of the OTA-FL are:
\begin{itemize}
    \item \textit{Efficiency:} OTA-FL can be more efficient in wireless environments, where wired connections might not be available or practical.
    \item \textit{Scalability:} It allows for the scalability of FL to a large number of devices, especially in settings where devices are mobile or have limited connectivity.
\end{itemize}

While FL has significant advantages, it presents several challenges including communication overhead, resource constraints, and deployment complexity \cite{wen2023survey}. Therefore, client selection has been introduced as a strategy to limit the number of communicating parties at every process step. 

\subsection{Client selection}
Client selection strategy involves determining which devices participate in each round of model training and contributing their updates to the central server to improve the FL quality and balance the need for privacy, data diversity, and model performance. However, randomly sampling clients in each training round may not fully exploit the local updates from heterogeneous clients, resulting in lower model accuracy, slower convergence rate, and degraded fairness \cite{fu2023client,smestad2023systematic}. Thus, selecting a client must consider various constraints for the following reasons: 
\begin{itemize}
    \item \textit{Resource Constraints:} Mobile devices participating in OTA-FL often have limited computational resources, battery life, and bandwidth. Client selection strategies need to take account of these constraints to minimize energy consumption, reduce latency, and optimize resource utilization.
    \item \textit{Dynamic network conditions:} OTA-FL operates in dynamic wireless environments where network conditions can vary significantly. Client selection algorithms must adapt to changes in connectivity, signal strength, and device availability to maintain efficient and reliable communication.
    \item \textit{Scalability and distributed computation:} As OTA-FL deployments scale to accommodate numerous devices and data sources, the client selection framework must be scalable and capable of distributed computation to handle the computational and communication overhead.
\end{itemize}
 
\subsection{Grey Wolf Optimizer}
The GWO is a metaheuristic optimization algorithm inspired by the social behavior of grey wolves in nature. It was introduced in 2014 \cite{mirjalili2014grey}. The algorithm simulates the grey wolves' leadership hierarchy and hunting mechanisms to solve optimization problems. The GWO is characterized by simulating the social hierarchy of a wolf pack with alpha, beta, delta, and omega wolves representing different solutions. It balances exploration and exploitation by assigning roles: the alpha wolf explores, while beta and delta wolves exploit. The omega wolf maintains diversity. Encouraging collaboration mimics real wolf pack cooperation, aiding in escaping local optima. Positions of wolves are iteratively updated based on movement equations inspired by hunting and social behaviors, contributing to the algorithm's convergence towards the global optimum in the search space.
The basic steps of the GWO can be summarized as follows:
\begin{enumerate}
    \item \textit{Initialization:} Initialize a population of wolves, representing potential solutions to the optimization problem.
    \item \textit{Objective function:} Evaluate the objective function for each wolf in the population.
    \item \textit{Update positions:} Update the positions of wolves based on the movement equations inspired by wolf behavior.
    \item \textit{Boundary handling:} Ensure that the updated positions of wolves remain within the defined search space.
    \item \textit{Selection:} Update the alpha, beta, and delta wolves based on their fitness values.
    \item \textit{Iteration:} Repeat the process until a stopping criterion is met such as a maximum number of iterations or a desired level of convergence.
\end{enumerate}

The GWO has been applied to various optimization problems in engineering, economics, and other fields \cite{faris2018grey,makhadmeh2023recent} due to its benefits, such as: 
\begin{itemize}
    \item \textit{Global Search Capability:} GWO exhibits a robust global search capability, making it well-suited for optimization tasks where finding a global optimum is crucial. This characteristic is particularly advantageous in complex problem spaces with multiple peaks.
    \item \textit{Fast Convergence Rate:} The algorithm is known for its fast convergence, allowing it to reach near-optimal solutions quickly. This is advantageous in scenarios where computational resources are limited, and rapid decision-making is required.
    \item \textit{Ease of Implementation:} GWO's simplicity facilitates straightforward implementation, making it accessible to practitioners with varying levels of expertise. The ease of implementation expedites the integration of GWO into diverse applications.
\end{itemize}
\section{Related Work and Our Contribution}
Recent advances in FL have focused on various aspects, including communication efficiency, privacy preservation, and robustness to adversarial attacks. However, the challenge of optimal client selection remains underexplored. Effective client selection is crucial for fast convergence, accurate models, fairness, and efficient communication. This section presents a literature review focused on optimizing client selection through various methods and highlights our contribution to this area.

{\renewcommand{\arraystretch}{1.5}%
\begin{table*}[!ht]

\centering
\begin{tabular}{|c|c|c|c|c|c|c|c|c|c|c|}
\hline
\textbf{Ref} & \textbf{Year} & \textbf{Accuracy} & \textbf{Energy} & \textbf{Delay} & \textbf{Reliability} & \textbf{Fairness} & \textbf{Model} & \textbf{Implementation} \\

 \hline \hline
 \cite{abdulrahman2020fedmccs}
 & 2020 & \checkmark & \checkmark & \checkmark & & & Dynamic programming & Traditional FL\\

 \cite{ruan2021valuable}
  & 2021 & \checkmark & \checkmark & \checkmark & & & Dynamic programming & Traditional FL\\
 
 \cite{zheng2021federated}
 & 2021 & \checkmark & \checkmark & \checkmark & & & Dynamic programming & Traditional FL\\

 \cite{zhang2023delay}
 & 2023 & \checkmark & \checkmark & \checkmark & & & Dynamic programming & Traditional FL\\
 
 \cite{huang2022stochastic}
 & 2022 & \checkmark & \checkmark & & & & Multi armed bandit & Traditional FL\\ 
 
 \cite{qu2022context}
 & 2022 & \checkmark & \checkmark & \checkmark & & & Multi armed bandit & Traditional FL\\

 \cite{zhu2022online}
 & 2023 & \checkmark & \checkmark & \checkmark & & \checkmark & Multi armed bandit & Traditional FL\\

\cite{huang2020efficiency} 
 & 2023 & \checkmark &  & \checkmark & & \checkmark & Multi armed bandit & Traditional FL\\

 \cite{shi2023efficient}
 & 2023 & \checkmark & \checkmark & \checkmark & &  & Multi armed bandit & Traditional FL\\
 
 \cite{kang2023ga}
  & 2023 & \checkmark & \checkmark & & & & Genetic algorithm & Traditional FL\\

 \cite{chahoud2023demand}
 & 2023 & \checkmark & \checkmark & & & \checkmark & Genetic algorithm & Traditional FL\\

 Our article
 & 2024 & \checkmark & \checkmark & \checkmark & \checkmark & \checkmark & Grey wolf optimizer & Over-the-air FL\\
 
 \hline
\end{tabular}
\vspace*{0.1cm}
\caption{related existing works on heuristic algorithm-based client selection}
\label{table:0}
\end{table*}}

\subsection{Random Selection}
This client selection method is achieved by randomly selecting a subset of clients to participate in the FL process. The work in \cite{nishio2019client} mitigates this problem and performs FL while actively managing clients based on their resource conditions by asking the randomly selected clients to send their resource information and participate in determining which of them go to complete the FL process. However, this approach presents several challenges such as building and maintaining client trust and ensuring high data quality. The random selection's implementation is simple but may lead to uneven data distribution and performance.

\subsection{Learning-based Selection}
Some papers implement client selection using ML techniques, where a central model predicts which clients provide high-quality updates. For instance, reinforcement learning is deployed to improve client selection performance by involving a reinforcement learning agent that learns a client selection policy \cite{cheng2023learning}. The authors in \cite{10.1145/3603166.3632563} introduced a clustering-based client selection framework to decrease the communication costs for training FL models by reducing the number of training devices at every round and the number of rounds required to reach convergence. Another learning-based client selection is proposed in \cite{10.1109/TC.2024.3355777} a comprehensive framework for client selection in FL based on the concept of value-of-information (VoI), which measures how valuable a client is for the global model aggregation, the VoI estimator uses reinforcement learning to learn the relationship between VoI and various heterogeneous factors of clients. The authors of \cite{wang2020optimizing} designed a framework that intelligently chooses the client devices to participate in each round to counterbalance the bias introduced by non-IID data and to speed up convergence. {A client sampling method was proposed in \cite{wu2023fedprof} to select relevant clients and mitigate the impact of low-quality data on the training process.}
Although this method allows for adaptive client selection strategies, it is computationally intensive, requires additional training, and may be sensitive to the quality of the initial model. 

\subsection{Heuristic Algorithm-based Selection}
Some methods formulate the client selection strategy as a mathematical optimization problem. Then, clients are selected using mathematical methods such as the dynamic programming model in \cite{zheng2021federated}, where the authors proposed a framework to balance the trade-off between the energy consumption of the edge clients and the learning accuracy of FL. The authors in \cite{abouzahir2023federated} proposed a predictive quality of service paradigm that allows devices to self-adjust their power allocation to maintain reliability and latency within the tolerated range of the URLLC application. In \cite{zhang2023delay}, the authors proposed a delay-constrained client selection framework for heterogeneous FL in intelligent transportation systems to improve the model performance such as accuracy, training, and transmission time. The multi-armed bandit (MAB) model is used in \cite{huang2022stochastic} to work for the hierarchical FL in MEC networks by estimating the participation probability for each client using the following information wireless channel state, local computing resources, and previous performance. The authors of \cite{qu2022context} also formulated the client selection problem as an MAB problem to design a selection framework where the network operator learns the number of successful participating clients to improve the training performance as well as under the limited budget on each edge server. \textbf{Contextual combinatorial MAB is used in \cite{shi2023efficient} to formulate a client selection problem to boost volatile FL by speeding up model convergence, promoting model accuracy, and reducing energy consumption.} The authors in \cite{zhu2022online} leveraged the MAB framework and the virtual queue technique in Lyapunov optimization to conduct client selection with a fairness guarantee in the asynchronous FL framework. \textbf{In \cite{huang2020efficiency}, it was found that fairness criteria play a critical role in the FL training process. A fairer client selection strategy can lead to higher final accuracy, though it may come at the cost of some training efficiency.}
Authors of \cite{kang2023ga} proposed a client selection method using a Genetic algorithm, which enables faster central model training at a lower cost based on the client’s cost and the result of its local update. A dynamic and multicriteria scheme for client selection is developed in \cite{chahoud2023demand} to offer more volume and heterogeneity of data in the FL process using a genetic algorithm. 

\subsection{Our contributions}
\subsubsection{\textbf{Multi-attribute client selection:}}
Based on related works (See Table \ref{table:0}), certain selection methods choose the clients with the best performance or high resources. This approach results in clients with low-level resource capacity being unable to participate in the training process, and their datasets being ignored. This leads to biased and unfair selection, which ultimately results in an underfitting of the learned global model for those low-level clients. Moreover, some proposed methods suffer from some futility of the clients which train their local models and then the server does not aggregate them. This leads to a waste of client energy. While existing works have primarily focused on accuracy and cost criteria for client selection, it is imperative to take into account other attributes such as reliability, fairness, privacy preservation, and energy efficiency. By incorporating these additional dimensions, we can foster more equitable participation, protect user privacy, and optimize resource utilization in OTA-FL systems, enhancing their overall performance and global model generalization.

\subsubsection{\textbf{Integration of GWO with client selection:}}
Given the scalability and efficiency requirements of OTA-FL, the GWO holds significant promise for optimizing client selection strategies. By leveraging GWO's global search capability and fast convergence rate, we can design client selection algorithms that effectively balance exploration and exploitation. Additionally, GWO's ease of implementation makes it well-suited for deployment in distributed environments with resource-constrained devices. Furthermore, GWO's ability to maintain diversity in the population of wolves can address the challenge of heterogeneity among FL clients. By ensuring that the client selection process considers diverse attributes and characteristics of participating devices, we can enhance the performance and robustness of OTA-FL systems.
As shown in Table \ref{table:0}, GWO has not been applied to optimize client selection to enhance FL model's accuracy, cost, energy efficiency, reliability, and fairness. This notable absence of GWO-based approaches in existing literature underscores a significant research gap and provides compelling motivation to explore its potential in this context. Our contributions are summarized below:
\begin{itemize}
    \item Offering a multi-attribute client selection framework that is noticed in the "select then train" method. It balances the accuracy with energy, delay, reliability, and fairness criteria to tackle the OTA-FL challenges such as security risks, limited computational capability, and unstable networks.
    \item Adopting the grey wolf algorithm to choose the set of eligible clients to join the learning process.
    \item Evaluating the proposed approach and analyzing the FL model performance in terms of accuracy, convergence time, and energy efficiency.
\end{itemize}

\section{Multi-Attribute Client Selection}
We consider an FL framework consisting of a single base station and $n$ clients $N = \{1, 2, \cdots, n$\}. Each client ${i}$ possesses local data, denoted as $D_{i}$. For each communication round, the server aims to learn a global model with the data $D_{i}$ distributed across the selected clients. 

To model the FL problem, we define the weight vector $w$ to capture the parameters related to the global model. We introduce the loss function $l(w, x_{j}, y_{j})$, which captures the FL performance over input vector $x_{j}$ and output $y_{j}$ for each $D_{i}$. The categorical cross-entropy is used as a loss function in performing the classification problem in our paper. The total loss function of client $i$ writes \cite{yang2020delay}:
\begin{equation}\label{Fi}
F_{i}(w) = \frac{1}{D_{i}} \sum_{j=1}^{D_{i}} l(w,x_{j},y_{j}).
\end{equation}
The FL training problem can be formulated as follows:
\begin{equation}\label{FL_model}
\min F(w) = \sum_{i=1}^{n} \frac{D_{i}}{D}F_i(w),
\end{equation}
where $D = \sum_{i=1}^{n} D_{i}$ is the total data samples of all clients.
\subsection{Delay}
To implement FL over wireless networks, wireless devices must train the model locally and transmit their results over wireless links. However, this computation and transmission introduce a delay that impacts the overall FL performance. Therefore, it is crucial to optimize the delay for efficient FL implementation.\\

\subsubsection{Computation Delay}
The computation delay is determined by the type of learning models and the desired learning accuracy $\epsilon_{i}$, the computation time at user $i$ needed for processing is \cite{yang2020delay}:
\begin{equation}\label{C_Delay}
\tau_{i}^{c} = \frac{C_{i} D_{i}}{f_{i}} \upsilon_{i} \log_2\left(\frac{1}{\epsilon_{i}}\right),
\end{equation}
where $\upsilon_{i}\log_2(1/\epsilon_{i})$ is the number of local iterations required for client $i$ to reach the desired accuracy $\epsilon_{i}$, $C_{i}$ (cycles/bit) is the number of CPU cycles required for computing one sample data at user $i$, and $f_{i}$ is the computation capacity of user $i$, which is measured by the number of CPU cycles per second.

{\renewcommand{\arraystretch}{1.2}%
\begin{table}[t]
\caption{Main notations used in this paper.}
\begin{tabular}{c|l}
 \textbf{Notation} & \textbf{Meaning} \\
  $n$ & Number of clients \\
  $D_{i}$  &  Data samples collected by client $i$ \\
  $D$      &  Total data samples of all users\\
  $w$      &  Global model parameter vector \\
  $x_{j}$  &  Input vector for each data sample $j$ \\
  $y_{j}$  &  Output vector for each data sample $j$ \\
  $l(w, x_{j}, y_{j})$  & Total loss function for client $i$\\
  $F_{i}(w)$ & Local objective function \\
  $F(w)$ & Global objective function \\
  $\tau_{i}^{c}$ & Computation delay \\
  $\tau_{i}^{t}$ & Transmission delay \\
  $\tau_{i}$ & Delay requirement \\
  $e_{i}^{c}$ & Computation energy \\
  $e_{i}^{t}$ & Transmission energy \\
  $e_{i}$ & Energy consumption requirement \\
  $\epsilon_{i}$ &  The desired learning accuracy \\
  $C_{i}$  &  Computation capacity required (CPU cycles per bit)\\
  $f_{i}$  &  Computation capacity of $i$ (CPU cycles per second) \\
  $r_{i}$  &  Transmission rate \\
  $b_{i}$  &  Bandwidth allocated to user $i$ \\
  $g_{i}$  &  Channel gain between user $i$ and the BS \\
  $p_{i}$  &  Transmit power of user $i$ \\
  $N_{0}$  &  Power spectral density of the Gaussian noise \\
  $M(w)$  &  FL model size\\
  $T$ & Total number of communication rounds \\
  $\zeta_{i}$  & Energy consumption factor of client $i$\\
  $MTBF_{i}$ & Mean time between failures\\
  $\rho_{i}(t)$ &  Reliability of client $i$ \\
  $\rho$ & Reliability requirement\\
  $m_{i}$ & Number of failures\\
  $c_{i}$ & Required minimum selection fraction for client $i$\\
  $\mathbf{X}_{p}$ & Position of the prey\\
  $\mathbf{X}$ & Position of the wolf\\
  $\mathbf{A}$ and $\mathbf{C}$ & GWO coefficient vectors\\
  $\mathbf{d}$  & Distance between the wolf and the prey\\
  $num\_clients$ & Number of selected clients\\
  
 \end{tabular}
\label{table:1}
\end{table}}
\subsubsection{Transmission Delay}
After local computation, all users upload their local FL parameters to the server, the quality of the wireless channel is the primary factor that determines the transmission rate in each round that is given by:
\begin{equation}\label{T_rate}
r_{i} = b_{i} \log_2\left(1+ \frac{g_{i} p_{i}}{N_{0}b_{i}}\right),
\end{equation}
where $b_{i}$ represents the bandwidth allocated to user $i$, $p_{i}$ is the transmit power of user $i$, $g_{i}$ is the channel gain between user $i$ and the BS, and $N_{0}$ is the power spectral density of the Gaussian noise. 

The model size determines the transmission time between the client and server, expressed as $M(w)$. The model transmission time is calculated using the following formula:
\begin{equation}\label{T_Delay}
\tau_{i}^{t} = \frac{M(w)}{r_{i}}.
\end{equation}

\subsection{Energy}
Energy is a critical factor to consider when deploying FL, to implement energy-efficient ML algorithms, optimize communications, use low-power hardware accelerators, and develop energy-aware scheduling strategies. Balancing the benefits of FL with the energy constraints of participating devices is crucial for its widespread adoption and long-term sustainability. The energy consumption of each client $i$ is the sum of the energy used to train the model on each client's device and the energy used to transmit the local model from the device to the server. 

\subsubsection{Computation Energy}
The computing resources consumed by model training depend on the size of local data $D_{i}$, which is expressed as \cite{chen2020joint}:
\begin{equation}\label{C_Energy}
e_{i}^{c} = \zeta_{i} f_{i}^2 \cdot \tau_{i}^{c} f_{i} = \zeta_{i} f_{i}^2 C_{i} D_{i} \upsilon_{i}  \log_2\left(\frac{1}{\epsilon_{i}}\right).
\end{equation}

where $\zeta_{i}$ is the energy consumption coefficient depending on the chip of each client $i$’s device. Note that, since the server has a continuous power supply, we do not consider the energy consumption of the server in our problem.

\subsubsection{Transmission Energy}
The energy consumption of client $i$ in model transmission is expressed as:
\begin{equation}\label{T_Energy}
e_{i}^{t} = p_{i} \tau_{i}^{t} = p_{i} \frac{M(w)}{r_{i}}.
 \end{equation}

\subsection{Reliability}
Choosing clients capable of completing local training is a crucial maintenance metric to measure performance, safety, and equipment design, especially for critical or complex assets. The reliability of the client's device ensures the trustworthiness, stability, and efficiency of the FL process. It allows FL systems to make informed decisions regarding the participation of clients, data quality, and security, which results in better model performance and a more dependable and robust learning process \cite{park2022federated}.
The reliability computation of a client $i$ is performed by considering the time between failures i.e. MTBF (mean time between failures), which refers to the average time between two failures and is defined as follows \cite{sharma2023reliable}:
\begin{equation}\label{MTBF}
MTBF_{i} = \frac{\tau_{i}^{c}}{m_{i}},
\end{equation}
where $m_{i}$ is the number of failures. The client reliability, or the probability of operating without fail for a time $t$, is denoted by $\rho_{i}(t)$:
\begin{equation}\label{reliability}
\rho_{i}(t) = e^{- t/ MTBF_{i}}.
\end{equation}
A higher reliable client device is less likely to fail shortly. This, in turn, reduces the risk of losing the training data or the local model due to unintentional shutdown and network instability.

\subsection{Fairness}
During the FL process, the client selection method often prioritizes devices with low latency. However, this bias towards speed may not be fair to clients with high data quality, the local dataset which has a larger size and whose distribution is more similar to the global distribution plays a more important role, and the corresponding clients should participate in more communication rounds. Therefore, it is important to consider the fairness constraint to avoid an overabundance of relevant clients \cite{smestad2023systematic}. The fairness constraint is considered to "tell" each client how many communication rounds they should participate \cite{xia2020multi}. We introduce the following constraint on a minimum selection fraction for each client $i$ \cite{li2019combinatorial}:
\begin{equation}\label{fairness}
\frac{1}{T} \sum_{t=1}^{T} E[a_{i}(t)] \geq c_{i},
\end{equation}
where $E[.]$ is the expectation operator and $c_{i} \in (0, 1)$ is the minimum fraction of communication rounds required to choose client $i$. $T$ is the total number of rounds and $a_{i}(t)$ is a binary variable defined as an indicator with  $a_{i}(t) = 1$ indicating that client $i$ is selected in round $t$, and $a_{i}(t) = 0$ otherwise.
\subsection{Problem Formulation}
Our approach involves a "select then train" client selection method where the server invites clients who meet the constraints of accuracy, energy, delay, reliability, and fairness to participate in the FL algorithm. We formulate our problem whose goal is to minimize the loss function of an FL algorithm by optimizing the various wireless parameters, as follows:
\begin{subequations}\label{eq:Server_Reward}
\begin{equation}
\min F(w) = \frac{1}{D}  \sum_{i=1}^{n} \sum_{j=1}^{D_{i}} l(w,x_{ji},y_{ji})
\tag{\ref{eq:Server_Reward}}
\end{equation}
    \begin{align}
    s.t. \quad \tau_{i}^{c} + \tau_{i}^{t} \leq \tau_{i}, \qquad\qquad \forall i \in N
    \label{eq:delay_condition}
    \end{align}
    \begin{align}
    \qquad 0 < e_{i}^{c} + e_{i}^{t} \leq e_{i}, \qquad \forall i \in N
    \label{eq:energy_condition}
    \end{align}
    \begin{align}
    \rho_{i}(t) \geq \rho, \qquad \quad \forall i \in N
    \label{eq:reliability_condition}
    \end{align}
    \begin{align}
     \frac{1}{T} \sum_{t=1}^{T} E[a_{i}(t)] \geq c_{i} \quad \forall i \in N
    \label{eq:fairness_condition}
    \end{align}
    \begin{align}
    \epsilon_{min} \leq \epsilon_{i} \leq 1 \quad \forall i \in N
    \label{eq:accuracy_condition}
    \end{align}
    \begin{align}
    0 \leq f_{i} \leq f_{i}^{max} \quad \forall i \in N
    \label{eq:iteration_condition}
    \end{align}
    \begin{align}
    0 \leq p_{i} \leq p_{i}^{max} \quad \forall i \in N
    \label{eq:tpower_condition}
    \end{align}
    \begin{align}
    \sum_{i=1}^{n} b_{i} \leq B \quad \forall i \in N
    \label{eq:bandwidth_condition}
    \end{align}
    \begin{align}
    0 \leq c_{i} \leq 1 \quad \forall i \in N
    \label{eq:f_condition}
    \end{align}
\end{subequations}
where $\gamma_{T}$ is the maximum delay to join the FL system, $\gamma_{E}$ is the energy consumption of the FL algorithm, $\gamma_{R}$ is the minimum reliability needed to participate to the FL process. 

Constraint (\ref{eq:delay_condition}) indicates that the execution time of the local tasks and transmission time for all clients should not exceed the maximum completion time for the whole FL algorithm. (\ref{eq:energy_condition}) is the energy consumption constraint to perform the learning task. Constraint (\ref{eq:reliability_condition}) is the client's device reliability condition for joining the FL algorithm. Constraint (\ref{eq:fairness_condition}) is the fairness constraint to participate in the FL algorithm. The local accuracy constraint is given by (\ref{eq:accuracy_condition}). Constraints (\ref{eq:iteration_condition}) and (\ref{eq:tpower_condition}) respectively represent the maximum local computation capacity and average transmit power limits of all clients. Due to the limited bandwidth of the system, we have (\ref{eq:bandwidth_condition}), where $B$ is the total bandwidth. Constraint (\ref{eq:f_condition}) is the fraction of communication rounds required to ensure a fair selection.

\section{Grey Wolf Optimizer-Based client selection}
\subsection{Federated Learning Algorithm}
Our FL system is depicted in the pseudo-algorithm \ref{alg:FL}. It is divided into two pieces, one executed by the server and the other by the clients. The server first initializes the global model parameters with random values. The server coordinates different rounds of execution. At each round, the server selects the set of clients using Algorithm \ref{alg:CS-GWO} and, in parallel, sends a copy of the training model. To fine-tune the copy of the training model, each client performs a series of gradient descent steps using its data. After training, each client sends back the weights and biases of the local model to the server. The server aggregates the updates from all clients and starts a new round.

\begin{algorithm}
\caption{OTA-FL with Multi-Attribute Client Selection }\label{alg:FL}
\begin{algorithmic}
\Statex
\textbf{Base Station Side:} 
\State Initialize the global model $W_{0}$ 
    \For{$t \gets 0$ to $T$}                    
        \State Select client set $\mathcal{C}$ using \textbf{Algorithm 2}
        \State Broadcast $W_{t}$ to selected clients (i.e., $\mathcal{C}$).
        \State Receive the over-the-air aggregated global model $W_{t+1}$.
    \EndFor

\State \textbf{Selected Client Side:}
\State At each round $t$:
\State Receive current global model $W_{t}$.
\State Train local model and produce model update $W_{t+1}^{c}$.
\State Send $W_{t+1}^{c}$ to the server.
\end{algorithmic}
\end{algorithm}

The number of selected clients is determined dynamically in each round based on several factors:
\begin{itemize}
    \item The number of clients available in each round.
    \item The total available bandwidth $B$: Each client's bandwidth requirement $b_i$ is considered to avoid exceeding $B$.
    \item The computation and energy: The computational power and energy availability of both the server and the clients are considered to avoid overburdening any participant.
\end{itemize}

\subsection{Client Selection Algorithm}
\begin{figure}[t]
\centering
\includegraphics[width=9cm]{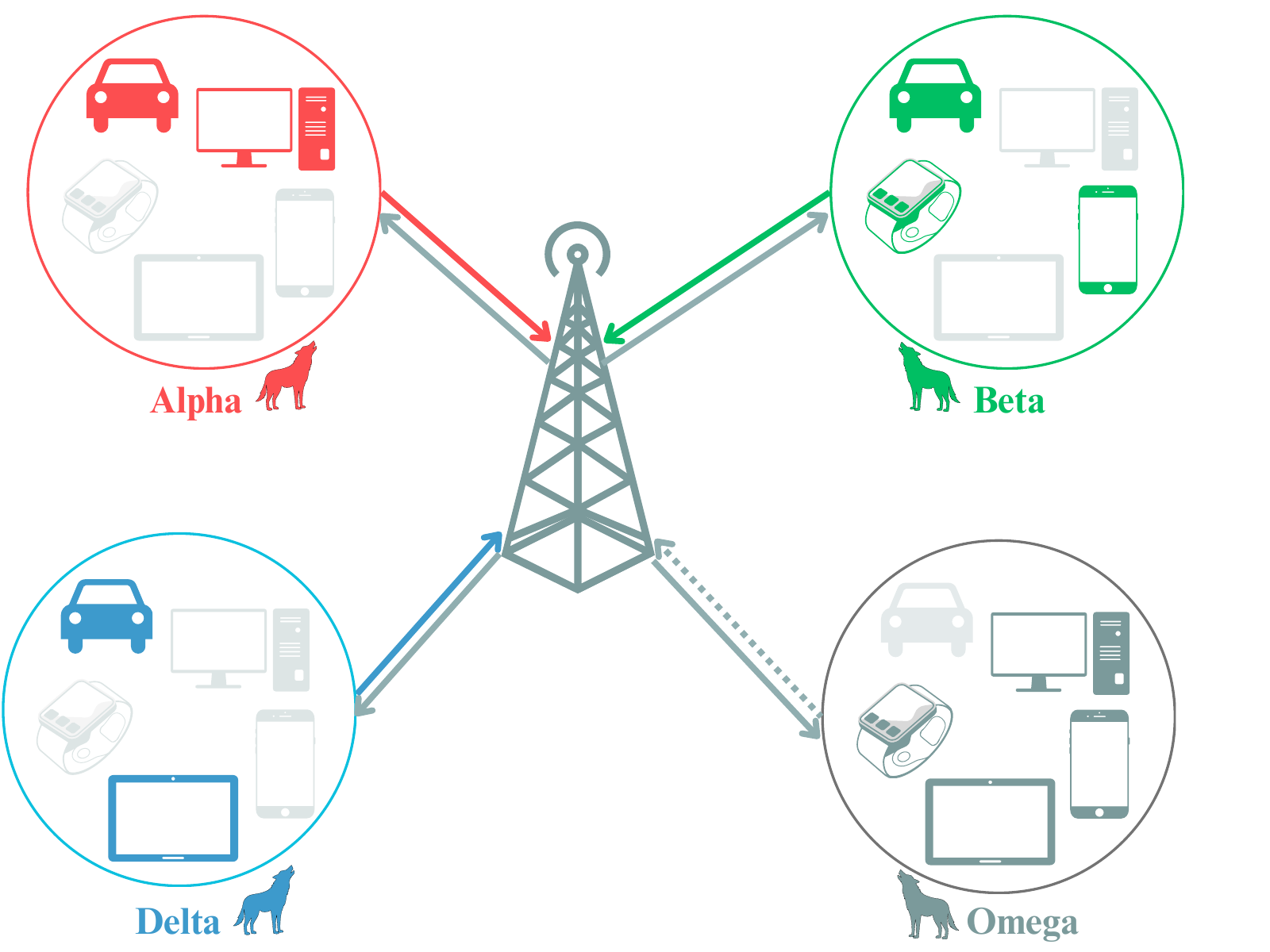}
\caption{The wolf in the GWO is the set of clients in the FL process: The selected clients are shown in bold pictures, while transparent pictures represent clients that have not been selected}
\Description{.}
\label{GWO}
\end{figure}

The GWO is a metaheuristic algorithm inspired by the social hierarchy and hunting behaviors of grey wolves in nature. It leverages these natural processes to efficiently search for optimal solutions in complex optimization problems due to the advantages of fewer parameters, simple principles, and implementation \cite{mirjalili2014grey}. In this work, we employ the grey wolf model for Optimizing the client selection problem (Eq. \ref{eq:Server_Reward}), wherein the wolf is represented as the set of clients that are eligible to join the learning process (See Fig.\ref{GWO}). 

Let’s assume that there are $S$ solutions (sets of clients) in the search space, GWO classifies these solutions based on the objective function (Eq.\ref{eq:Server_Reward}) for four categories as follows: the best solution is alpha ($\alpha$), the second-best is beta ($\beta$), the third-best delta ($\delta$) and the rest solutions are omega ($\omega$). The best three solutions $(\alpha,\beta,\delta)$ are used to guide the other solutions $(\omega)$ for improving the search space. During the optimization, there are three main phases of hunting behavior: Encircling, hunting, and attacking which will be detailed later.

\subsubsection{Encircling Phase}
The grey wolves start hunting by creating a circle around the prey. The mathematical model of the encircling phase is developed using the following equations:
\begin{equation}\label{encyrcling}
    \mathbf{X}(t+1)=\mathbf{X}_{p}(t) - \mathbf{A}\times\mathbf{d}.
\end{equation}
The distance $\mathbf{d}$ between the wolf and the prey is calculated by the following equation:
\begin{equation}\label{Distance}
    \mathbf{d} = |\mathbf{C}\times\mathbf{X}_{p}(t) - \mathbf{X}(t)|,
\end{equation}
where $t$ is the current iteration, $\mathbf{X}_{p}$ is the position of the prey and $\mathbf{X}$ is the position of the wolf. $\mathbf{A}$ and $\mathbf{C}$ are coefficient vectors defined as follows:
\begin{align}
\mathbf{A} &= 2\mathbf{a}\times\mathbf{r}_{1}-\mathbf{a},\label{A}\\
\mathbf{C} &= 2\mathbf{r}_{2}.\label{C}
\end{align}
The components of $\mathbf{a}$ are linearly decreased from 2 to 0 over iterations and can be calculated by:
\begin{equation}\label{a}
a = 2 - t \times 2/max_{itr},
\end{equation}
where $max_{itr}$ is the maximum number of iterations. $\mathbf{r}_{1}$ and $\mathbf{r}_{2}$ are random vectors in $[0,1]$.

\subsubsection{Hunting Phase}
During the hunting phase, the three most promising solutions denoted by $(\alpha,\beta,\delta)$ are obtained. As for the other research agents $(\omega)$, they need to update their positions by moving towards the average of the three best-known positions since they have better knowledge about the optimal location of the prey. In this regard, the following equations have been presented with $i\in\{\alpha, \beta, \delta\}$:
\begin{equation}\label{Xalpha}
\mathbf{X}_{i}(t+1) = \mathbf{X}_{i}(t) - \mathbf{a}_{i} \times \mathbf{d}_{i},
\end{equation}
where $\mathbf{d}_{i}$ is estimated using the following:
\begin{equation}\label{Distance_alpha}
    \mathbf{d}_{i} = \left|\mathbf{C}_{i}\times\mathbf{X}_{i}(t) - \mathbf{X}(t)\right|.
\end{equation}
Let $p_{i}$ be the positive weight associated with wolf $i\in\{\alpha, \beta, \delta\}$ such that $\sum_{i}p_{i}=1$. Given the positions of wolves $\alpha, \beta$, and $\delta$, a good estimation of the average position of the optimal solution at round $t$ is given by:
\begin{equation}\label{Xnext iteration}
\mathbf{X}(t+1) = \sum_{i\in\{\alpha,  \beta, \delta\}}p_{i}\cdot \mathbf{X}_{i}(t+1).
\end{equation}

\subsubsection{Attacking Phase}
GWO finishes hunting by attacking the prey when it stops moving, to model approaching the prey we use Eq. (\ref{a}) as the parameter $a$ is responsible for making the balance between exploration and exploitation, the value of a linearly decreased from 2 to 0 over iterations, consequently, the parameter $A$ takes a random value in the interval $[-2a, 2a]$ given by Eq. (\ref{A}). The wolves take a random position when $A > 1$ or $A < -1$ and are forced to move towards the prey when $-1 \leq A \leq 1$.

\begin{algorithm}
\caption{Grey Wolf Optimizer-Based client Selection}
\label{alg:CS-GWO}
\begin{algorithmic}
\Statex

Initialize the grey wolf population $\mathbf{X}$ \\
Initialize a, A, and C \\
Calculate the fitness of each search agent \\
$X_{\alpha}$ = the best search agent \\
$X_{\beta}$ = the second best agent \\
$X_{\delta}$ = the third best search agent
\While{$t<max_{itr}$}
    \For{each search agent} 
        \State Randomly initialize $r_1$ and $r_2$ 
        \State Update the position of the current search agent using Eq.(\ref{Xnext iteration})
    \EndFor    
    \State Update a, A, and C
    \State Calculate the fitness of all search agents
    \State Update $X_{\alpha}, X_{\beta}$, and $X_{\delta}$
    \State $ t = t+1$
\EndWhile
\State return $X_{\alpha}$ \Comment{Best solution: Set of clients to join the FL}
\end{algorithmic}
\end{algorithm}

The multi-attribute client selection is provided in Algorithm.\ref{alg:CS-GWO}. First, the GWO parameters are initialized by the base station by randomly setting the positions of wolves within the defined problem bounds, ensuring diversity in the initial population. The $X_\alpha$, $X_\beta$, and $X_\delta$ wolves, representing the best solutions found, are initially set to zero vectors and updated as the algorithm progresses. The coefficient $a$ decreases linearly from 2 to 0 over the iterations, balancing exploration and exploitation. $A$ and $C$ used in the position update formulas are derived from $a$ and random values $r1$ and $r2$.
Second, the GWO calculates the score of the best clients based on the lowest loss value, lowest computation and transmission delay, lowest energy consumption, highest reliability, and fairness. The best score value is sent to the BS from each set of clients. The algorithm tracks the best positions for the $X_\alpha$, $X_\beta$, and $X_\delta$ wolves based on their fitness, updating these whenever a better solution is discovered. We set $max_itr$ to 50, consistent with common practice in the literature, to ensure effective exploration and exploitation of the search space. Finally, the local models are trained by the best clients with the best score $X_\alpha$ and sent to the base station for aggregation via OTA communication.
\section{Experimental Investigation}
To assess the effectiveness of the proposed multi-attribute client selection algorithm for FL systems, we conducted experiments to analyze the performance of the global model and investigate the effects of delay, energy consumption, reliability, and fairness constraints. In this section, We offer a comparative analysis between our solution and several existing methods, including dynamic programming, multi-armed bandit, and genetic algorithms. Our evaluation will be based on various datasets including MNIST, CIFAR-10 and Fashion MNIST, considering test loss, test accuracy, energy consumption, training time, reliability, and fairness as key metrics.

\subsection{Experimental setup}
Under our problem formulation, the "select then train" method is effectively implemented using client metadata and historical performance data. Although the objective function is dependent on the model parameters and local datasets, the selection process leverages surrogate metrics derived from client profiles, which include computational capabilities, data size, and distribution summaries. These profiles are updated periodically and shared with the server, allowing it to make informed decisions without direct access to the local data. We implement our FL model using the following datasets:
\begin{itemize}
    \item MNIST, comprising 60,000 28 x 28 images of handwritten digits from 0 to 9.
    \item CIFAR-10, which includes 60,000 32 x 32 color images in 10 classes, with 6,000 images per class.
        \item Fashion MNIST, consisting of 70,000 28 x 28 grayscale images of 10 different categories of clothing items.
\end{itemize}
These datasets are distributed among 50 clients to train the FL model, and each client possesses unique hardware parameters, metadata, and statistics. This includes information such as data size, historical performance, device capabilities, network conditions, and previous training outcomes. This setup enables us to compute the total delay using Eq. (\ref{C_Delay}) and Eq. (\ref{T_Delay}), the total energy using Eq. (\ref{C_Energy}) and Eq. (\ref{T_Energy}), reliability using Eq. (\ref{reliability}), and fairness using Eq. (\ref{fairness}). 

Our experimentation took place on the Google Colab T4 GPU cloud-based platform, utilizing Python version 3, TensorFlow version 2.3.0, and Keras version 2.4.3 for code development. We employed the convolutional neural network (CNN) algorithm to tackle our classification problems, employing the stochastic gradient descent (SGD) technique for training acceleration. Our study aims to enhance FL performance using GWO (Algorithm \ref{alg:CS-GWO}) by selecting clients capable of achieving optimal scores in prediction accuracy, delay, energy consumption, reliability, and fairness.

\subsection{Comparison Scheme}
In our previous work \cite{driss2024gwo}, we compared random client selection, loss-aware client selection, and our multi-attribute client selection using the MNIST dataset. This comparison demonstrated the effectiveness of our multi-attribute method in achieving superior performance metrics compared to simpler client selection techniques. The experimental results indicate that the proposed multi-attribute client selection can reduce energy consumption by up to 43\% compared to the random client selection method. Additionally, our multi-attribute method outperforms the loss-aware method in terms of time reduction, computational efficiency, and energy consumption. These ablation experiments demonstrate that a comprehensive approach considering various attributes for client selection such as delay, energy, fairness, and reliability is more effective for FL systems.

\renewcommand{\arraystretch}{1.3}%
\begin{table}
\caption{Global accuracy and energy efficiency under different client selection methods.}
\begin{center}
\begin{tabular}{|c|c|c|} 
 \hline
 \textbf{Client Selection Method} & \textbf{Accuracy (\%)} & \textbf{Energy Efficiency (\%/joule)}\\ 
 \hline
 Random selection (3 clients) & 68 & 0.46\\
 \hline
 Random selection (5 clients) & 73 & 0.41\\
 \hline
 Loss-aware client selection & 92 & 0.80\\ 
 \hline
 Our multi-attribute client selection & \textbf{98} & \textbf{1.08}\\
 \hline
\end{tabular}
\label{FL_accuracy}
\end{center}
\end{table}
To further validate our approach, we conduct experiments on larger datasets with a larger number of clients. We compare the proposed solution with the following approaches:
\begin{itemize}
    \item \textbf{Dynamic programming (DP):} is a classic algorithm to solve the knapsack problem, an optimization problem that involves selecting a subset of items from a given set, each with a weight and a value. The objective is to maximize the total value of the chosen items while ensuring that the cumulative weight does not surpass a specified capacity \cite{assi2018survey}.
    \item \textbf{Genetic algorithm (GA):} is an evolutionary optimization technique inspired by the process of selection and genetics. It mimics natural evolution, where individuals with higher fitness are more likely to survive and reproduce, leading to the emergence of better solutions over generations \cite{thengade2012genetic}.
    \item \textbf{Multi-armed bandit (MAB):} is a classic problem in probability theory and decision-making, often used in the context of optimization and resource allocation. The name originates from the idea of a gambler facing multiple slot machines (the "bandits"), each with potentially different payoff probabilities, and need to decide which machine to play to maximize their total reward over time \cite{burtini2015survey}. We implemented the upper confidence bound (UCB) algorithm, which is designed to address this trade-off by selecting options based on a combination of their average rewards and the uncertainty or confidence interval around those rewards. The UCB algorithm helps balance exploration and exploitation by assigning a score to each option, which includes an upper bound term to encourage exploring less-tried options \cite{ottens2017duct}.
\end{itemize}

\subsection{Experimental Results}
To demonstrate the efficiency of our client selection approach in OTA-FL, we analyze the FL model using the MNIST, CIFAR10, and Fashion MNIST classification problems. This analysis aimed to evaluate various performance metrics, including the global model accuracy, loss probability, convergence time, energy consumption, energy efficiency, reliability, and fairness. Furthermore, we conducted a comparative assessment, juxtaposing the outcomes of our multi-attribute client selection employing GWO against those of other established methods, such as dynamic programming, multi-armed bandit, and genetic algorithms.

\renewcommand{\arraystretch}{1.4}%
\begin{table}[h!]
    \centering
    \caption{Global performance of our OTA-FL system under different client selection schemes}
    \label{tab:comparison_all}
    \begin{tabular}{|p{3.2cm}||p{2.0cm}|p{1.6cm}|p{1.2cm}|p{1.8cm}|p{1.1cm}|p{1.5cm}|}
    \hline
        \textbf{Multi-attribute client selection method} & \textbf{Dataset} & \textbf{Accuracy (\%)} & \textbf{Loss Probability} & \textbf{Convergence Time (s)} & \textbf{Total Energy (J)} & \textbf{Energy Efficiency (\%/J)} \\
        \hline \hline
        \multirow{3}{*}{Genetic algorithm} & MNIST & \hfil96.31 & \hfil0.0505 & \hfil15415 & \hfil12000 & \hfil0.008 \\
        \cline{2-7}
        & CIFAR-10 & \hfil72.25 & \hfil0.045 & \hfil26960 & \hfil26100 & \hfil0.0027 \\
        \cline{2-7}
        & Fashion MNIST & \hfil82.73 & \hfil0.034 & \hfil22600 & \hfil23400 & \hfil0.0035 \\
        \hline
        \multirow{3}{*}{Multi-armed bandit} & MNIST & \hfil98.15 & \hfil0.0185 & \hfil15029 & \hfil12800 & \hfil0.0076 \\
        \cline{2-7}
        & CIFAR-10 & \hfil75.68 & \hfil0.040 & \hfil25200 & \hfil27760 & \hfil0.0027 \\
        \cline{2-7}
        & Fashion MNIST & \hfil85.36 & \hfil0.032 & \hfil21300 & \hfil24700 & \hfil0.0034 \\
        \hline
        \multirow{3}{*}{Dynamic programming} & MNIST & \hfil98.07 & \hfil0.0193 & \hfil11422 & \hfil11920 & \hfil0.0082 \\
        \cline{2-7}
        & CIFAR-10 & \hfil75.15 & \hfil0.043 & \hfil24340 & \hfil26350 & \hfil0.0028 \\
        \cline{2-7}
        & Fashion MNIST & \hfil84.49 & \hfil0.033 & \hfil20600 & \hfil23600 & \hfil0.0035 \\
        \hline
        \multirow{3}{*}{Grey wolf optimizer} & MNIST & \hfil\textbf{98.43} & \hfil\textbf{0.0173} & \hfil\textbf{11200} & \hfil\textbf{11800} & \hfil\textbf{0.0084} \\
        \cline{2-7}
        & CIFAR-10 & \hfil\textbf{77.78} & \hfil\textbf{0.039} & \hfil\textbf{23100} & \hfil\textbf{24500} & \hfil\textbf{0.0031} \\
        \cline{2-7}
        & Fashion MNIST & \hfil\textbf{86.25} & \hfil\textbf{0.031} & \hfil\textbf{19500} & \hfil\textbf{22300} & \hfil\textbf{0.0044} \\
        \hline
    \end{tabular}
    \Description{Comparison}
\end{table}

\begin{figure*}
\centering
\subfigure[15 clients.]{\includegraphics[width=0.49\linewidth]{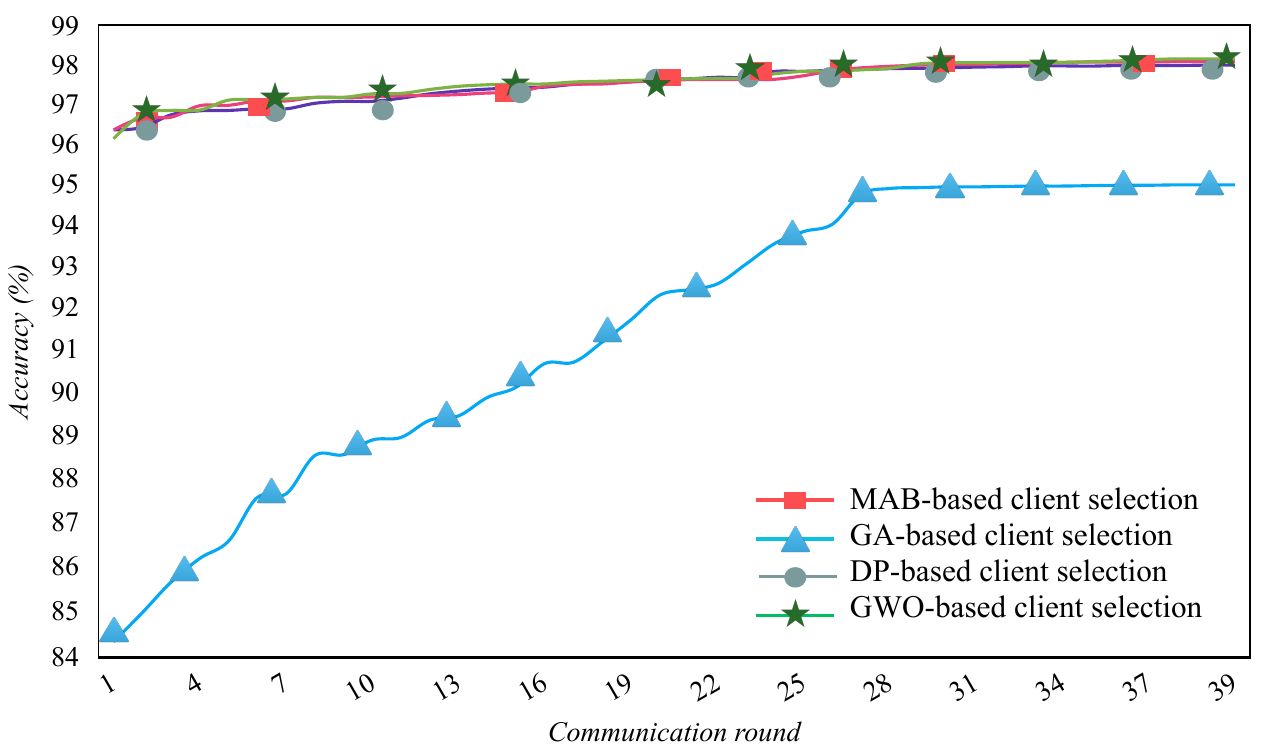}}
\subfigure[50 clients.]{\includegraphics[width=0.49\linewidth]{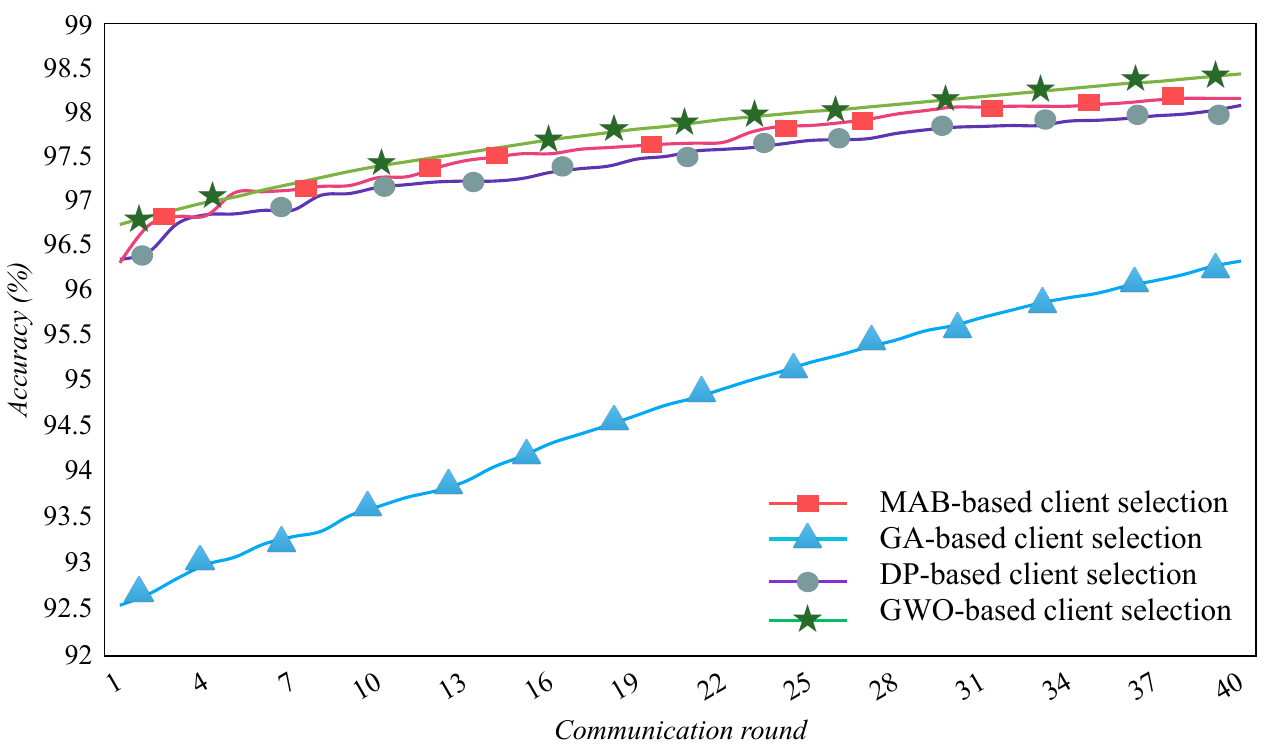}}
\caption{Test accuracy under different client selection schemes for MNIST dataset.}
\label{acc_epochs}
\Description{Comparison}
\end{figure*}

\noindent\textbf{Accuracy:}
The accuracy metric indicates the overall performance of the FL model across all clients. It represents the proportion of correctly predicted instances in the entire dataset. In our FL system, we observe in (Table \ref{tab:comparison_all}) and (Fig.\ref{acc_epochs}) that the client selection approach using the GWO and MAB achieved the highest accuracy. This indicates that the selection strategy based on the grey wolf algorithm and multi-armed bandit effectively utilized client resources and data contributions to improve model accuracy. The other two methods, using the GA and DP algorithms, also achieved high accuracy. While slightly lower than the top-performing methods, this accuracy still indicates robust performance in accurately predicting image labels.

\begin{figure}[!ht]
  \subfigure[15 clients.]{\includegraphics[width=0.49\linewidth]{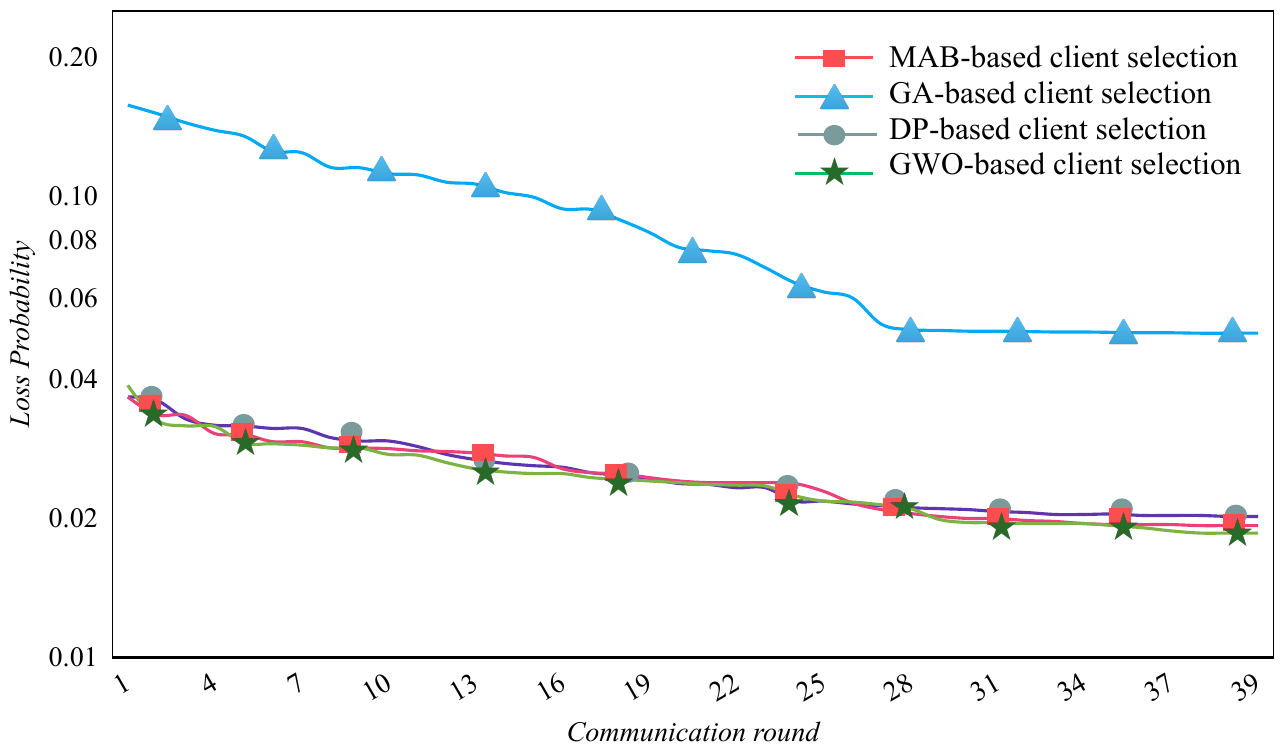}}\label{fig:sub-3}
\subfigure[50 clients.]{\includegraphics[width=0.49\linewidth]{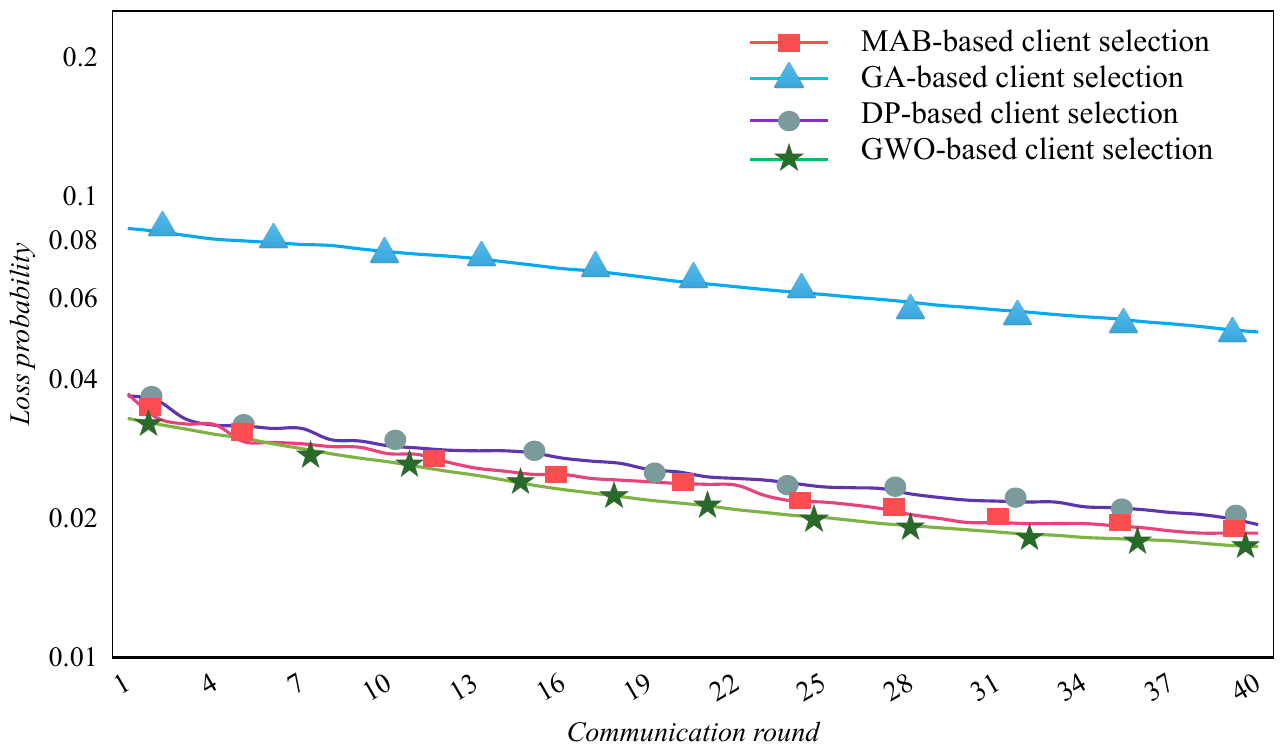}}\label{fig:sub-4}
  \caption{Loss probability under different client selection schemes for MNIST dataset.}
  \Description{.}
  \label{Loss}
\end{figure}

\noindent\textbf{Loss Probability:}
The loss probability offers valuable insight into the uncertainty surrounding model predictions, reflecting the likelihood of erroneous predictions or misclassifications. In our analysis, employing client selection methods such as the MAB algorithm and the GWO resulted in remarkably low loss probabilities (See Table \ref{tab:comparison_all}). This achievement underscores a high degree of confidence in the accuracy of their predictions. As depicted in (Fig. \ref{Loss}), it's evident that the loss probability consistently decreased across various iterations using the MNIST dataset, further affirming the effectiveness and reliability of the applied methodologies.\\

\begin{figure}
  \subfigure[15 clients.]{\includegraphics[width=0.49\linewidth]{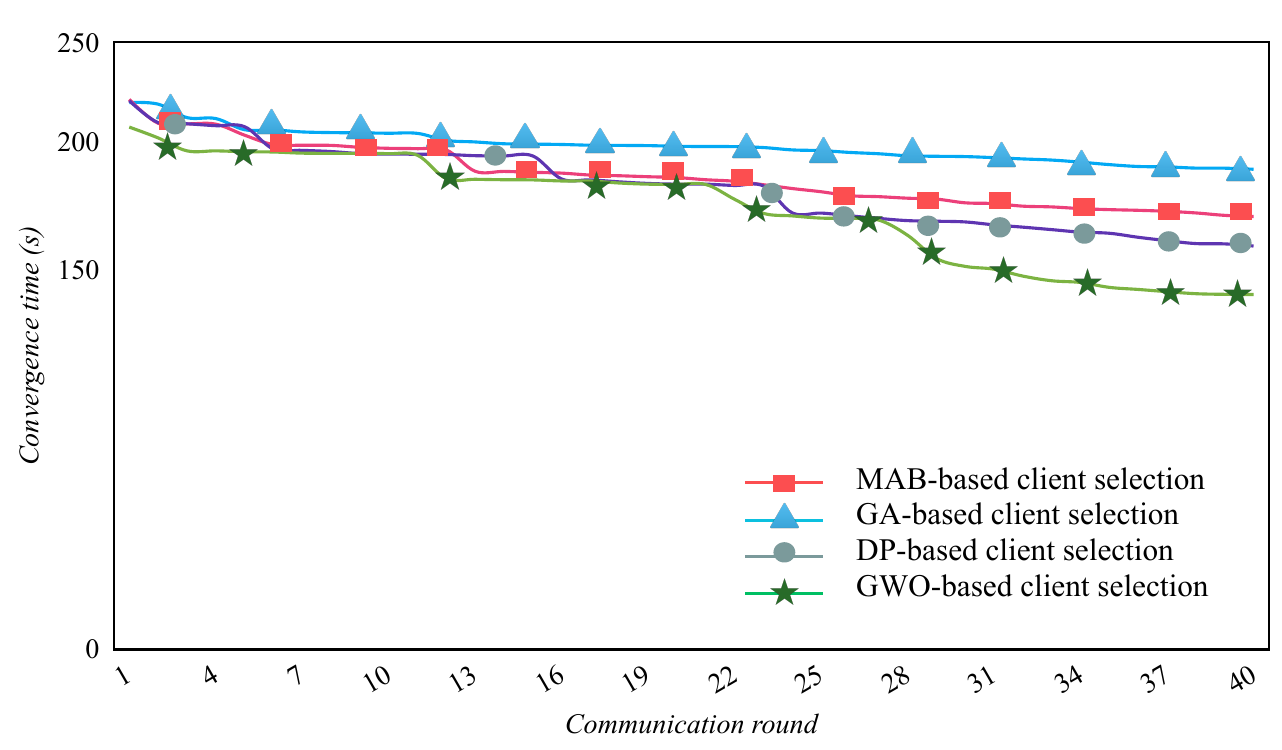}}\label{fig:sub-5}
\subfigure[50 clients.]{\includegraphics[width=0.49\linewidth]{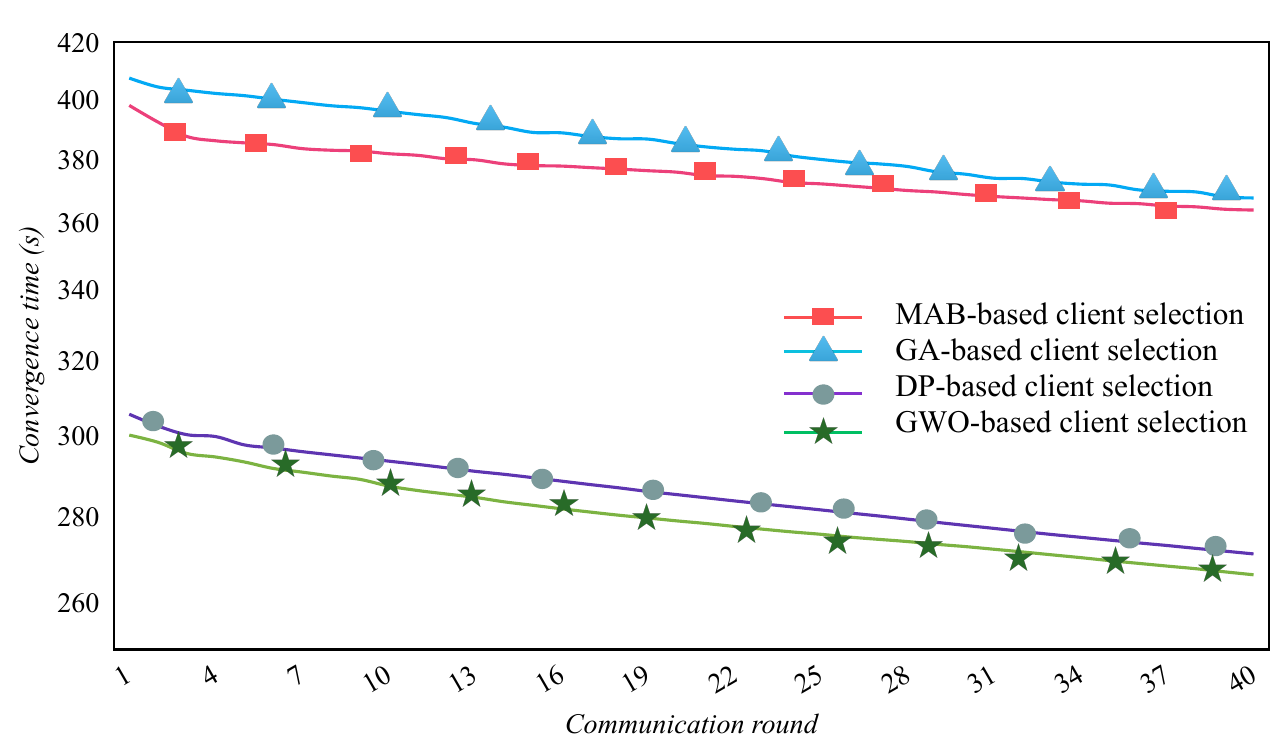}}\label{fig:sub-6}
  \caption{Convergence time under different client selection schemes for MNIST dataset.}
  \Description{.}
  \label{time_method}
\end{figure}

\noindent\textbf{Convergence Time:}
The GWO exhibits a gradual decrease in execution time (as seen in Fig.\ref{time_method}), indicating efficient convergence towards optimal client selections over epochs, with stabilization observed after reaching a steady state. Similarly, the Multi-armed Bandit approach demonstrates steady execution time reduction, reflecting efficient convergence and balanced exploration-exploitation strategies. In contrast, the dynamic programming and genetic algorithms show more variability in execution times, with fluctuations observed throughout epochs, suggesting differing convergence behaviors and computational requirements. The convergence time (See Table \ref{tab:comparison_all}) across different client selection methods reveals notable differences in computational efficiency. Our solution based on the GWO emerges as the most time-efficient method, with a total execution time of 6925 seconds. This indicates that the GWO-based multi-attribute client selection approach requires the least amount of time to complete the FL process compared to the other methods.

\noindent\textbf{Energy Efficiency:}
The energy consumption across different client selection methods in our FL system reveals notable variations in energy efficiency. As seen in Table \ref{tab:comparison_all}, the GWO-based multi-attribute client selection consistently demonstrates the lowest energy consumption, maintaining a constant energy level for all communication rounds. This suggests that the GWO-based client selection approach maintains energy efficiency by consistently selecting optimal clients without significant fluctuations. In comparison, both the DP and GA-based approaches exhibit higher energy consumption across all epochs. Similarly, the multi-armed bandit approach also shows consistent energy consumption. These results indicate that the GWO-based approach is the most energy-efficient among the methods evaluated, highlighting its potential for reducing energy costs in our OTA-FL system.

\noindent \textit{Instantaneous Energy Efficiency (IEE):} To assess the energy efficiency at round $t$ of our scheme and analyze how it compares with existing literature, we introduce the following energy efficiency indicator:
\begin{equation}\label{EE_t}
    IEE(t)  \triangleq \frac{\mbox{Accuracy(t)}}{\mbox{Energy(t)}} \quad\mbox{(\%/joule)}.
\end{equation}

\noindent \textit{Global Energy Efficiency (GEE):} Similarly, we assess the global energy efficiency of our schemes using the following energy efficiency indicator :
\begin{equation}\label{EE}
    GEE  \triangleq \frac{\mbox{Global Accuracy}}{\mbox{Total Energy}} = \frac{\mbox{Accuracy(T)}}{\sum\limits_{t=1}^{T}\mbox{Energy(t)}} \quad\mbox{(\%/joule)}.
\end{equation}

\begin{figure}
\subfigure[15 clients.]{\includegraphics[width=0.49\linewidth]{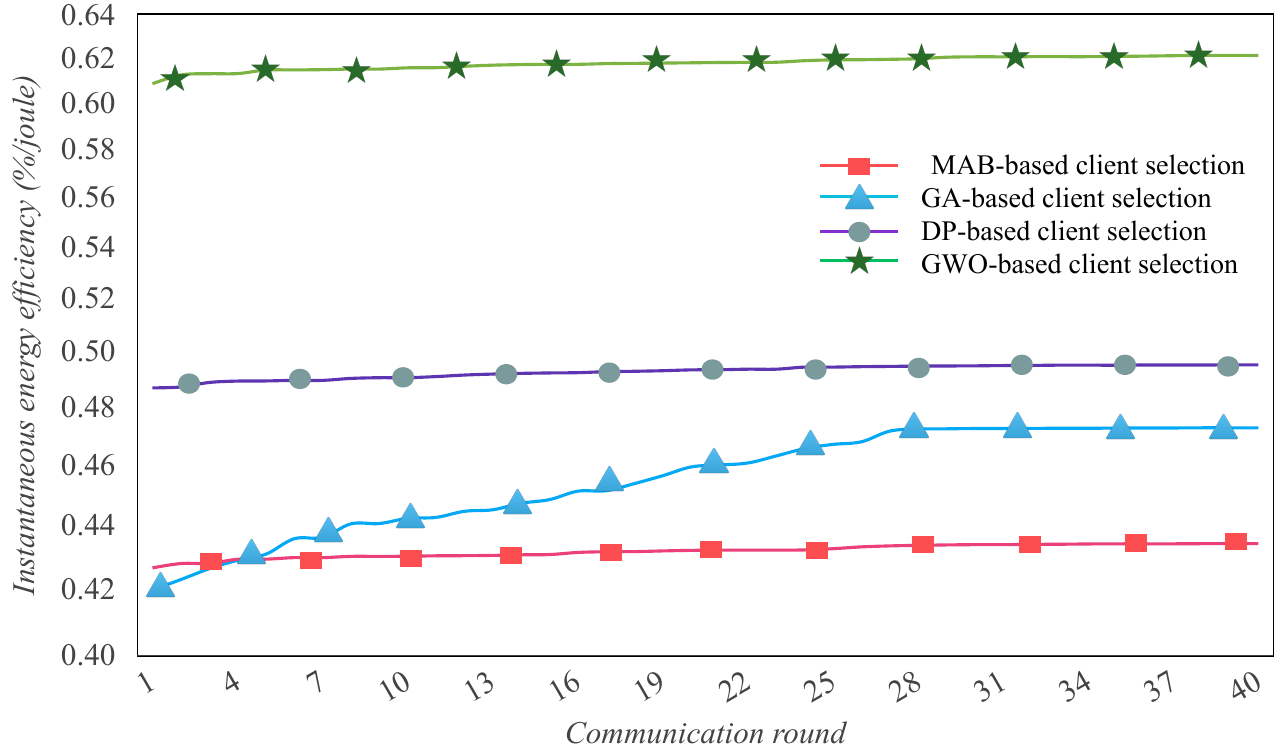}}\label{fig:sub-1}
\subfigure[50 clients.]{\includegraphics[width=0.49\linewidth]{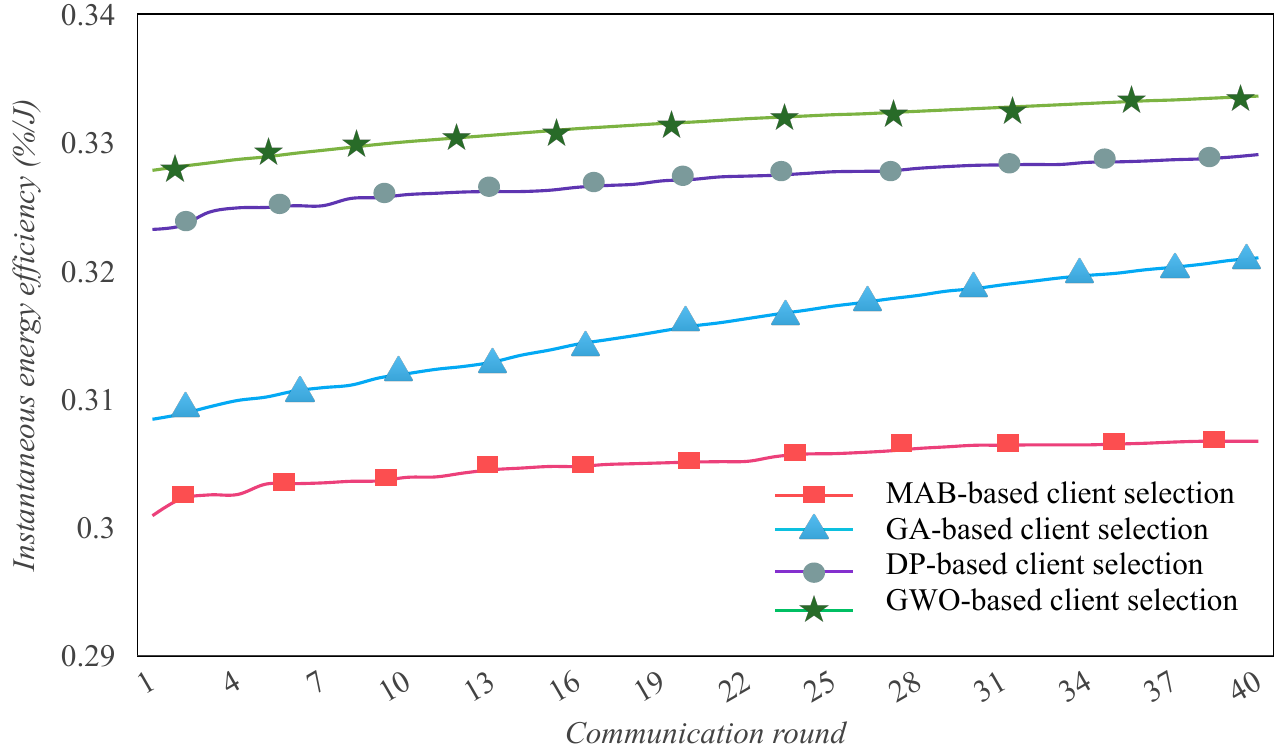}}\label{fig:sub-2}
  \caption{Instantaneous energy efficiency under different client selection schemes for MNIST dataset.}
  \Description{.}
  \label{EE_CS}
\end{figure}

\noindent The evolution of this instantaneous energy efficiency indicator under different client selection schemes is illustrated in Fig.\ref{EE_CS} which reveals several insights. Firstly, examining our multi-attribute client selection using the GWO, it's observed that the IEE metric consistently increases over communication rounds, suggesting a stable and potentially improving energy performance. This indicates that the GWO-based method optimizes client selection with minimal energy consumption. In contrast, the selection using the DP algorithm demonstrates slightly lower but relatively stable IEE values across iterations, indicating moderate energy efficiency. The genetic algorithm approach shows a gradual increase in IEE over communication rounds, suggesting a less stable but still moderate energy performance. Finally, the multi-armed bandit approach displays the lowest IEE values, with fluctuations across iterations, indicating relatively poor energy efficiency compared to other methods. In general, the GWO-based method stands out as the most energy-efficient approach with an accuracy of 0.0084\% per joule consumed for the MNIST dataset, followed by the DP and GA methods while the multi-armed bandit-based method appears to be the least energy-efficient in this context.

\noindent\textbf{Reliability:} 
The reliability metric in our FL system is crucial for ensuring the trustworthiness, stability, and efficiency of the FL process, especially when dealing with critical or complex assets. It measures the ability of clients to complete local training tasks reliably over time, thereby impacting performance, safety, and equipment design. The results illustrated in (Table \ref{Reliability_fairness_perf}) indicate that our multi-attribute client selection using the GWO achieves the highest average reliability of 0.70, followed by the dynamic programming, the genetic algorithm, and the multi-armed bandit method with the lowest average reliability. Additionally, examining the worst reliability values reveals that the GWO-based method also outperforms other methods with the worst reliability of 0.42. These results imply that the GWO method not only achieves higher average reliability but also maintains better worst-case reliability compared to other client selection methods in our system, indicating its superiority in selecting reliable devices for participation in the OTA-FL process.

\renewcommand{\arraystretch}{1.3}%
\begin{table} 
\centering 
\caption{Average and worst reliability and Fairness under different client selection schemes for MNIST dataset.}
\label{Reliability_fairness_perf} 
\begin{tabular}{|p{3cm}||p{2cm}|p{2cm}|p{2cm}|p{2cm}|} 
\hline 
\multirow{2}{*}{\parbox{3cm}{\textbf{Multi-attribute client}\\\textbf{selection method}}} & \multicolumn{2}{c|}{\textbf{Reliability}} & \multicolumn{2}{c|}{\textbf{Fairness}} \\
\cline{2-5} 
& \hfil \textbf{Average value} & \hfil \textbf{Worst value} & \hfil \textbf{Average value} & \hfil \textbf{Worst value} \\
\hline\hline 
Multi-armed bandit & \hfil 0.53 & \hfil 0.34 & \hfil \textbf{0.92} & \hfil \textbf{0.68} \\
\hline 
Genetic algorithm & \hfil 0.55 & \hfil 0.32 & \hfil 0.84 & \hfil 0.60 \\
\hline 
Dynamic programming & \hfil 0.60 & \hfil 0.36 & \hfil 0.86 & \hfil 0.64 \\
\hline 
Grey wolf optimizer & \hfil \textbf{0.70} & \hfil \textbf{0.42} & \hfil 0.89 & \hfil 0.65 \\
\hline 
\end{tabular} 
\end{table}

\noindent\textbf{Fairness:}
Our fairness constraint aims to ensure equitable treatment of all participating clients regarding their contribution to communication rounds. The results presented in Table \ref{Reliability_fairness_perf} highlight the strong emphasis on fairness criteria across all four multi-attribute client selection methods in our OTA-FL system, with average fairness scores ranging from 0.89 to 0.92. Our multi-attribute client selection using the multi-armed bandit achieves the highest average fairness score of 0.92, indicating effective balancing of client participation in communication rounds and ensuring a fair distribution of workload among them. Following closely is the GWO-based method, which also demonstrates a commendable performance in maintaining fairness. The GA and DP-based methods, although slightly lower in average fairness scores, still exhibit satisfactory performance in most cases, showcasing the overall robustness of our fairness constraint across different client selection techniques.

\renewcommand{\arraystretch}{1.6}%
\begin{table}[!ht]
\caption{Global performance of our OTA-FL system under two selection methods for MNIST dataset.}
\begin{center}
\begin{tabular}{|p{3.5cm}||p{1.8cm}|p{1.3cm}|p{1.6cm}|p{1.8cm}|p{2.5cm}|} 
 \hline
 \textbf{Our multi-attribute Client Selection Method} & \textbf{Convergence Time (s)} & \textbf{Accuracy (\%)} & \textbf{Loss\newline Probability} & \textbf{Total Energy (joule)} & \textbf{Energy Efficiency (\%/joule)}\\ 
 \hline\hline
 Train then Select & \hfil 14000 & \hfil 98.00 & \hfil 0.0180 & \hfil 12564 & \hfil0.0078\\
 \hline
 Select then Train & \hfil11200 & \hfil98.43  & \hfil0.0173 & \hfil11800 & \hfil0.0084 \\ 
 \hline
\end{tabular}
\Description{select then train results}
\label{PERF_ST_TS}
\end{center}
\end{table}
\noindent\textbf{Select then train VS Train then select:}
There are two primary methods for client selection strategy: selecting clients before training which aims to optimize the selection process before initiating model training, potentially reducing communication overhead, and ensuring that only the most important clients participate in the learning process. The other method is selecting clients after training local models which involves training local models on all available clients first and then selecting a subset of them based on their performance or contribution to the global model. The results presented in Table \ref{PERF_ST_TS} indicate that selecting clients before training the global model leads to faster convergence, higher accuracy, lower loss probability, and improved energy efficiency compared to training local models and then selecting clients. Additionally, this pre-selection of FL participants preserves the clients' privacy and enhances security by minimizing the opportunity for unauthorized access to client data during the selection process. As a result, the "Select then Train" method not only improves performance but also reinforces the confidentiality and integrity of client data, making it a more robust and privacy-preserving approach in FL environments.

\subsection{System scalability}
\renewcommand{\arraystretch}{1.3}%
\begin{table}
\centering
\caption{Global Performance of our approach under different numbers of clients.}
\label{Grey_Wolf_Performance}
\begin{tabular}{|p{2cm}||p{2cm}|p{2cm}|p{2.5cm}|p{2cm}|p{2cm}|}
\hline
\textbf{Number of Clients} & \textbf{Accuracy (\%)} & \textbf{Convergence time (s)} & \textbf{Energy Efficiency (\%/joule)} & \textbf{Average Reliability} & \textbf{Average Fairness}\\
\hline\hline
15 Clients & \hfil98.16 & \hfil6925 & \hfil0.016 & \hfil0.67 & \hfil0.87 \\
\hline
50 Clients & \hfil98.43 & \hfil19500 & \hfil0.0044 & \hfil0.70 &  \hfil0.89\\
\hline
\end{tabular}
\end{table}
To analyze the scalability of our multi-attribute client selection using the GWO, we evaluate performance metrics with varying numbers of clients. As the number of clients increases, accuracy improves (Fig.\ref{acc_epochs}), indicating that the model benefits from learning from a more diverse set of client data, which enhances generalization. This improvement demonstrates that more clients contribute valuable and varied data, leading to a better-performing model.
The energy efficiency shows significant improvement to indicate that while more clients require more resources, the utilization of these resources becomes more effective (Fig.\ref{EE_CS}). However, convergence time rises significantly with more clients (Fig.\ref{time_method}). This increase is primarily due to the additional communication overhead and computational complexity associated with aggregating updates from a larger number of clients. Furthermore, the reliability and fairness of the model improve with the addition of more clients demonstrating that a larger number of clients leads to a more stable and equitable distribution of resources and benefits. In summary, scaling up the number of clients enhances model accuracy, energy efficiency, reliability, and fairness at the price of increased complexity. In our daily lives, most applications require acceptable accuracy which may need a low-average number of clients keeping the complexity at acceptable levels.

\subsection{Discussion \& Insights}
By leveraging the multi-attribute client selection using GWO, we aim to optimize the process of choosing clients based on multiple attributes crucial to the success of OTA-FL. One key aspect is the ability of our solution to enhance the selection of participants based on their proficiency in providing informative updates. In OTA-FL, the quality of model updates plays a pivotal role in the overall learning process. Clients capable of contributing insightful and relevant updates contribute significantly to the effectiveness and generalization of the global model (see Table \ref{tab:comparison_all}). The GWO-based approach helps us identify and prioritize clients with a higher potential for delivering informative contributions, thereby enriching the learning experience. Moreover, the GWO assists in striking a balance between the informative updates and the associated communication costs. 

\renewcommand{\arraystretch}{1.6}%
\begin{center}  
\begin{table*}
\caption{Exploration, exploitation, and complexity of each client selection method.}
\label{complex}
\begin{tabular}{| p{2.7cm} || p{0.6cm} | p{8cm} | p{2.2cm} |} 
\hline  
\textbf{Client Selection Method} & \textbf{Ref} & \textbf{Exploration vs. Exploitation} & \textbf{Complexity} \\ \hline\hline  

Multi-attribute MAB-based client selection & \cite{mannor2004sample} & Balancing exploration (trying different arms to learn their rewards) and exploitation (favoring clients with higher expected rewards based on past observations) & \hfil $\mathcal{O}\left(n\cdot T\cdot\log{}n\right)$ \\ \hline 

Multi-attribute GA-based client selection & \cite{nopiah2010time} & Balancing exploration (diversity in clients) and exploitation (selecting and refining promising sets of clients through crossover and mutation) & \hfil $\mathcal{O}\left(n\cdot T\right)$\\ \hline 

Multi-attribute DP-based client selection & \cite{ezugwu2019comparative} & Focusing on exploitation, as the goal is to find the best combination of clients given the energy constraint & \hfil $\mathcal{O}\left(n\cdot T\cdot E\_dim\right)$ \\ \hline   
Multi-attribute GWO-based client selection & \cite{mirjalili2014grey} & Balancing exploration (searching for new promising sets of clients) and exploitation (exploiting known promising sets) through the movement of wolves towards better solutions &  \hfil $\mathcal{O}\left(n\cdot T\right)$ \\ \hline
\end{tabular} 
\end{table*}
\end{center} 
Table \ref{complex} shows that each multi-attribute client selection method balances exploration and exploitation differently to optimize performance. The MAB-based method focuses on exploring various client combinations while favoring those with higher expected rewards, with computational complexity scaling logarithmically with the number of selected clients ($num\_clients$). GA-based selection emphasizes diversity and refinement of promising solutions through crossover and mutation, exhibiting linear complexity with selected clients and communication rounds. The DP-based selection primarily exploits the best client combination within resource constraints, its complexity tied to the required energy consumption dimensionality ($E\_dim$). At the same time, the GWO-based method combines exploration and exploitation by iteratively refining client selections, with a complexity similar to the GA-based method. 

Analyzing the robustness of our multi-attribute client selection approach to factors such as noise, outliers, and changes in network conditions provides valuable insights into its reliability and resilience in real-world scenarios. Noise in client data, arising from measurement errors or inconsistencies, can potentially impact the performance of client selection algorithms by introducing inaccuracies or biases. For future work, we aim to evaluate the ability of this approach to handle noisy data effectively, either by incorporating noise-reduction techniques or by adapting selection criteria to account for variability in data quality. 
 
\section{Conclusion}
We proposed a multi-attribute client selection framework utilizing the GWO to strategically manage the number of participants in each round and enhance the OTA-FL process. Our framework effectively optimizes several critical factors, including accuracy, energy consumption, delay, reliability, and fairness of participating devices. Experimental results validate the robustness and scalability of our approach. Compared to state-of-the-art methods, our framework ensures that our FL system is not only more accurate and fair but also significantly more energy-efficient and responsive. This makes our approach particularly well-suited to meet the demands of modern applications, where efficient and equitable system performance is essential.

\bibliographystyle{ACM-Reference-Format}
\bibliography{journal_clean}

\end{document}